# Workspace Analysis of the Parallel Module of the VERNE Machine

Daniel Kanaan, Philippe Wenger, and Damien Chablat

Institut de Recherche en Communications et Cybernétique de Nantes, U.M.R. C.N.R.S. 6597
1, rue de la Noë, BP 92101, 44321 Nantes Cedex 03 France

*Abstract— The paper addresses geometric aspects of a spatial three-degree-of-freedom parallel module, which is the parallel module of a hybrid serial-parallel 5-axis machine tool. This parallel module consists of a moving platform that is connected to a fixed base by three non-identical legs. Each leg is made up of one prismatic and two pairs of spherical joint, which are connected in a way that the combined effects of the three legs lead to an over-constrained mechanism with complex motion. This motion is defined as a simultaneous combination of rotation and translation. A method for computing the complete workspace of the VERNE parallel module for various tool lengths is presented. An algorithm describing this method is also introduced.*

## I. INTRODUCTION

The workspace calculation of a parallel manipulator is very important for the designer and for the end-user. If we consider a serial robot, the representation of the workspace is generally based on the illustration in 3 dimensions of the space reachable by the center of its wrist (characterizing translations) and by the space reachable by the extremity of the terminal link (characterizing orientations), these two zones being uncoupled. Unfortunately, it is not the case for parallel robots: the zone reachable by the center of the moving platform is dependent on the orientation of its platform. Thus a graphical representation of the workspace of parallel manipulators with more than three degrees of freedom is only possible if we fix parameters representing the exceeded degrees of freedom. As consequence, different types of workspace were used in the literature, according to the choice of the presented parameters [1].

Several methods may be used to calculate the workspace of a parallel manipulator. One can mostly distinguish between discretization methods, geometrical methods, and analytical methods. A simple way for determining the workspace of a parallel manipulator is to use a discretization method. In this method, a grid of nodes with position and orientation is defined. Then each node is tested to see whether it belongs to the workspace or not [2, 3]. The discretization algorithm takes into account all constraints and it is simple to implement but is has some serious drawbacks. It is expensive in computational time and the accuracy depends on the sampling step that is used to create the grid [4]. Geometrical methods are mostly used to determine the boundary of the workspace. The principle is to define geometrical models for the constraints that limit the workspace of the parallel manipulator [5]. These models are obtained for each leg separately and the workspace is the intersection between these models [1]. Analytical methods are more difficult to apply because they increase the dimension of the problem by introducing supplementary variables. They consist in solving an optimization problem with penalties at the borders [6].

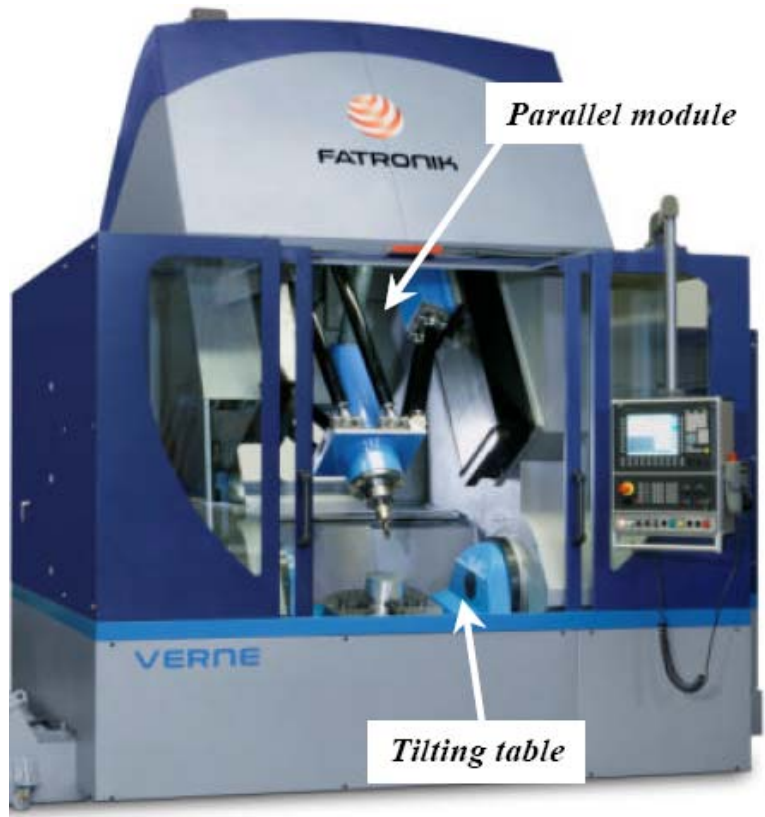

Figure 1: Overall View of the VERNE Machine

Parallel kinematic machines (PKM) are commonly claimed to offer several advantages over their serial counterparts [7], such as high structural rigidity, better payload-to-weight ratio, high dynamic capacities and high accuracy [1, 8]. Thus, they are prudently considered as promising alternatives for high-speed machining and have gained essential attention of a number of companies and researchers. Since the first prototype presented in 1994 during the IMTS in Chicago by Gidding and Lewis (the VARIAX) [9], many other parallel manipulators have appeared. However, most of the existing PKM still suffer from a limited range of motion [10]. This drawback can be diminished by designing a hybrid manipulator as for the VERNE machine, which is a 5-axis machine-tool built by Fatronik for IRCCyN [11]. This machine-tool consists of a parallel module and a tilting table as shown in Fig. 1. The parallel module moves the spindle mostly in translation while the tilting table is used to rotate the workpiece about two orthogonal axes.

A simplified workspace model of the Verne machine is used currently, but this model is reduced with respect to the real one. The purpose of this paper is to calculate the real workspace to enhance the working capability and to improve the productivity of the VERNE machine. In the following section, we present the VERNE parallel module and we formulate its geometric equations. Section III is devoted to the calculation of the complete workspace for various tool lengths of the VERNE parallel module. In this section we define geometric models for constraints limiting the workspace. Then we apply a combination of geometric and discretization methods in order to calculate the complete workspace. Finally a conclusion is given in section IV.

## II. DESCRIPTION OF THE VERNE PARALLEL MODULE

### A. *Parallel manipulator structure*

Figure 2 shows a scheme of the parallel module of the VERNE machine. The vertices of the moving platform are connected to a fixed-base plate through three legs I, II and III. Each leg uses pairs of rods linking a prismatic joint to the moving platform through two pairs of spherical joints. Legs II and III are two identical parallelograms. Leg I differs from the other two legs in that $A_{11}A_{12} \neq B_{11}B_{12}$, that is, it is not an articulated parallelogram. The movement of the moving platform is generated by three sliding actuators along three vertical guideways.

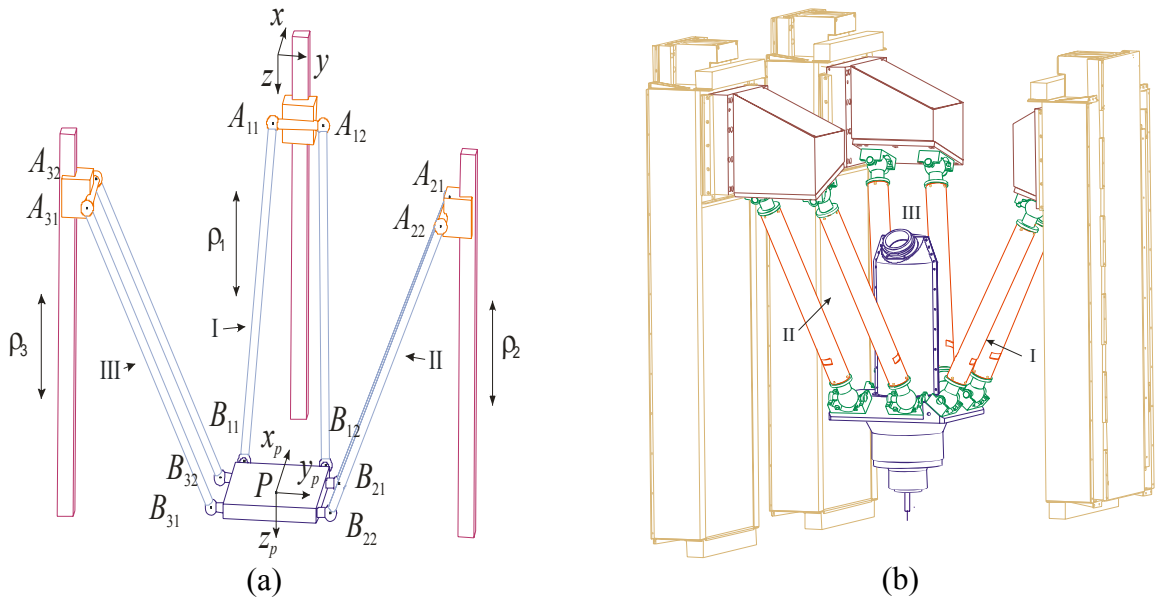

(a)                                                    (b)

Figure 2: Schematic representation of the Parallel Module; (a) simplified representation and (b) the real representation supplied by Fatronik

Due to the arrangement of the links and joints, as shown in Fig. 2, legs II and III prevent the platform from rotating about y and z axes. Leg I prevents the platform from rotating about z-axis but, because $A_{11}A_{12} \neq B_{11}B_{12}$, a slight coupled rotation about x-axis exists.

### B. Kinematic equations

In order to analyze the kinematics of the parallel module, two relative coordinates are assigned as shown in Fig. 2. A static Cartesian frame xyz is fixed at the base of the machine tool, with the z-axis pointing downward along the vertical direction. The mobile Cartesian frame, $x_P y_P z_P$, is attached to the moving platform at point P and remains parallel to xyz.

In any constrained mechanical system, joints connecting bodies restrict their relative motion and impose constraints on the generalized coordinates, geometric constraints are then formulated as algebraic expressions involving generalized coordinates.

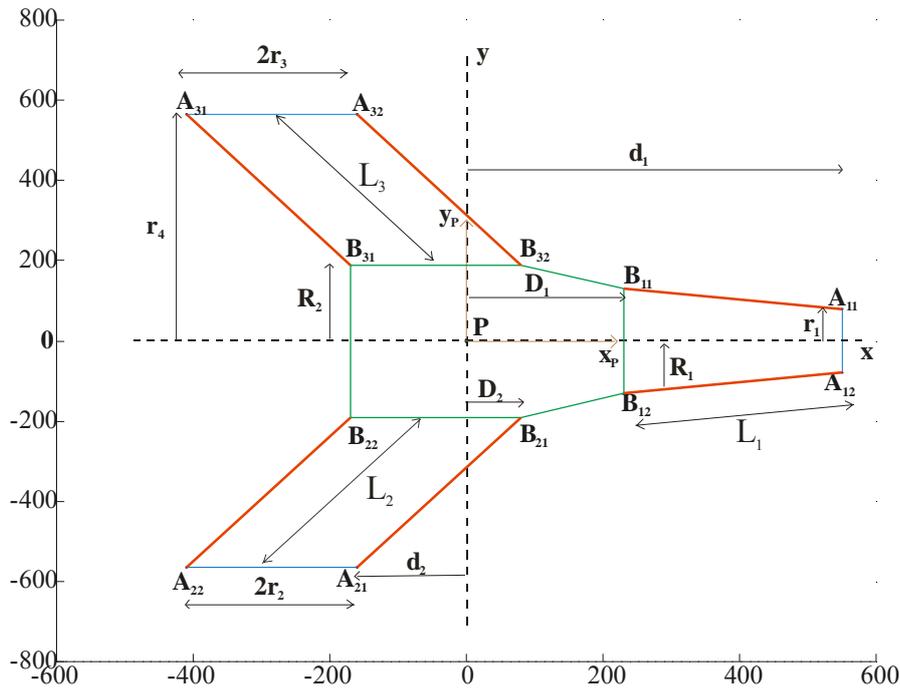

Figure 3: Dimensions of the parallel kinematic structure in the frame supplied by Fatronik

Using the parameters defined in Fig. 3, the constraint equations of the parallel manipulator are expressed as:

$$\left(x_{Bij} - x_{Aij}\right)^2 + \left(y_{Bij} - y_{Aij}\right)^2 + \left(z_{Bij} - z_{Aij}\right)^2 - L_i^2 = 0 \tag{1}$$

where $A_{ij}$ (respectively $B_{ij}$) is the center of spherical joint number j on the prismatic joint number i (respectively on the moving platform side), i = 1..3, j = 1..2.

Leg I is represented by two distinct equations defined by Eqs. (2a) and (2b). This is due to the fact that $A_{11}A_{12} \neq B_{11}B_{12}$ (Figure 3)

$$\left(x_P + D_1 - d_1\right)^2 + \left(y_P + R_1 \cos(\alpha) - r_1\right)^2 + \left(z_P + R_1 \sin(\alpha) - \rho_1\right)^2 - L_1^2 = 0 \tag{2a}$$

$$\left(x_P + D_1 - d_1\right)^2 + \left(y_P - R_1 \cos(\alpha) + r_1\right)^2 + \left(z_P - R_1 \sin(\alpha) - \rho_1\right)^2 - L_1^2 = 0 \tag{2b}$$

Leg II is represented by Eq. (3).

$$\left(x_P + D_2 - d_2\right)^2 + \left(y_P - R_2 \cos(\alpha) + r_4\right)^2 + \left(z_P - R_2 \sin(\alpha) - \rho_2\right)^2 - L_2^2 = 0 \tag{3}$$

The leg III, which is similar to leg II (Figure 3), is represented by Eq. (4).

$$\left(x_P + D_2 - d_2\right)^2 + \left(y_P + R_2 \cos(\alpha) - r_4\right)^2 + \left(z_P + R_2 \sin(\alpha) - \rho_3\right)^2 - L_3^2 = 0 \tag{4}$$

### C. Coupling between the position and the orientation of the platform

The parallel module of the VERNE machine possesses three actuators and three degrees of freedom. However, there is a coupling between the position and the orientation angle of the platform. The object of this subsection is to study the coupling constraint imposed by leg I.

By eliminating $\rho_1$ from Eqs. (2a) and (2b), we obtain a relation (5) between $x_P$, $y_P$ and $\alpha$ independently of $z_P$.

$$R_1^2 \sin(\alpha)^2 \left(x_P + D_1 - d_1\right)^2 + \left(r_1^2 - 2R_1 r_1 \cos(\alpha) + R_1^2\right) y_P^{\,2}$$
$$-R_1^2 \sin(\alpha)^2 \left(L_1^2 - \left(R_1^2 + r_1^2 - 2R_1 r_1 \cos(\alpha)\right)\right) = 0 \tag{5}$$

We notice that for a given $\alpha$, Eq. (5) represents an ellipse (6). The size of this ellipse is determined by $a$ and $b$, where $a$ is the length of the semi major axis and $b$ is the length of the semi minor axis.

$$\frac{\left(x_P + D_1 - d_1\right)^2}{a^2} + \frac{y_P^{\,2}}{b^2} = 1 \tag{6}$$

where
$$\begin{cases} a = \sqrt{\left(L_1^2 - \left(R_1^2 + r_1^2 - 2R_1 r_1 \cos(\alpha)\right)\right)} \\ b = \sqrt{\dfrac{R_1^2 \sin(\alpha)^2 \left(L_1^2 - \left(R_1^2 + r_1^2 - 2R_1 r_1 \cos(\alpha)\right)\right)}{\left(r_1^2 - 2R_1 r_1 \cos(\alpha) + R_1^2\right)}} \end{cases}$$

These ellipses define the locus of points reachable with the same orientation $\alpha$.

### III. Workspace Calculation of the VERNE Parallel Module

#### A. Preliminaries

The parallel architecture of the VERNE possesses 3 degrees of freedom; a complete representation of the workspace is a volume. Consequently the workspace of the VERNE machine can be defined by the positions in space reachable by a specific point connected to the moving platform.

However, since the first leg is not a parallelogram, the platform undergoes a parasitic rotation movement about the x-axis defined by the angle $\alpha$. This movement poses a problem in the determination of the workspace of VERNE, because it does not represent a controlled degree of freedom. Thus, it can be expressed as function of the coordinates of the point P ($x_P$, $y_P$ and $z_P$), centre of the mobile Cartesian frame.

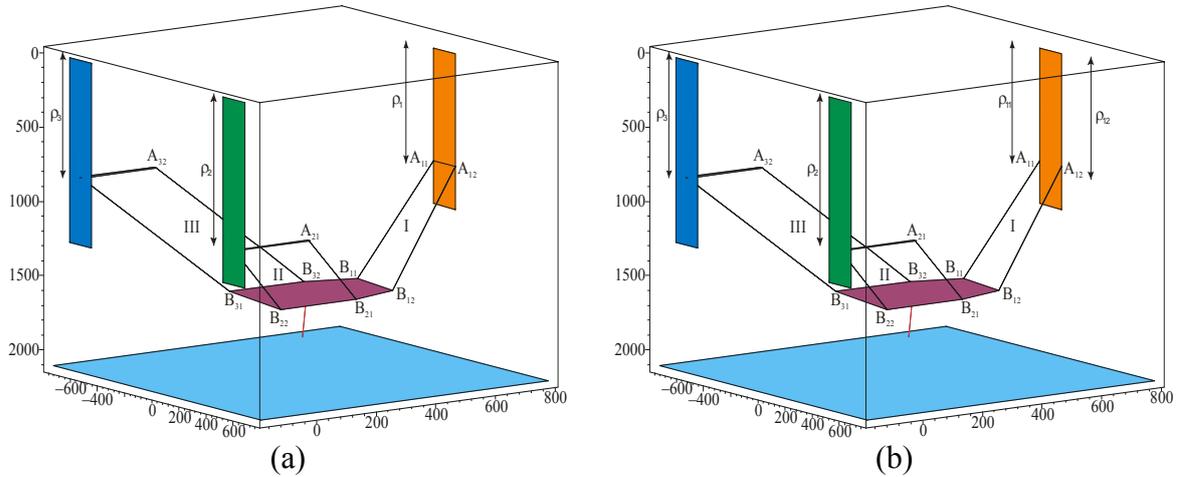

Figure 4: VERNE. (a) Real, (b) With virtual division of the leg I

To solve the problem created by the presence of the orientation $\alpha$, we propose the following method:

*Step 1: We virtually cut the leg I which is constituted of rods 11 and 12 (rod ij denotes $A_{ij}B_{ij}$) by supposing that $\rho_{11}$ is independent of $\rho_{12}$ (see Figure 4b). The parallel architecture of the VERNE possesses now 4 degrees of freedom instead of 3. These degrees of freedom are defined by coordinates $x_P$, $y_P$ and $z_P$ of the point P and by the orientation $\alpha$ of the moving platform. Thus we consider that the orientation $\alpha$ of the platform is given, and we geometrically model constraints limiting the workspace of the new parallel architecture (subsections C, D and E). The intersection between these models is a volume.*

*Step 2: We consider the interdependence between rods 11 and 12 of the leg I characterized by the fact that $\rho_{11} = \rho_{12} = \rho_1$ (Figure 4a). This will allows us to determine a relation between $x_P$, $y_P$, $z_P$ and $\alpha$. Then for a given orientation $\alpha$, the point P describes a surface (subsection F).*

*Step 3: The intersection between geometric models defined at step 1 and the surface defined at step 2 represents the constant orientation workspace of the VERNE. We calculate a horizontal cut of this workspace when the point P, center of the mobile Cartesian frame, moves in a known horizontal plane. Then we proceed by discretization to determine the complete workspace of the VERNE (subsection G).*

It is important to mention that the constraints imposed by the two opposite main rods that constitute the same leg possess the same limits on the workspace, if the leg has a shape of a parallelogram. Then it is sufficient to study one of these two rods to calculate the workspace. So taking into account this remark, we can limit the calculation of the workspace by studying only rods 11, 12, 21 and 31, since legs II and III have a shape of a parallelogram.

### B. Geometric models for constraints limiting the workspace

In this section, we virtually cut the leg I supposing that $\rho_{11}$ can be different from $\rho_{12}$ (Figure 4b) and we calculate for a given orientation $\alpha$, the workspace under the constraints on rods. This workspace is a volume because by cutting leg I we add one degree of freedom to the parallel architecture.

Generally, to calculate geometrically the workspace of a parallel manipulator, it is necessary to establish geometric models for all the constraints limiting the workspace [12]. In our case, we only take into account limitations on rods, and we try to determine geometrically the boundary of the workspace under the hypothesis that the constraints on the rods permit to define the maximal region that point $B_{ij}$ can cross, where $B_{ij}$ is the point of attachment of the rod ij to the platform, this in an independent manner for every rod.

Suppose that the constraints on rod ij allow us to define the reachable volume $V_{ij}$ by point $B_{ij}$. When $B_{ij}$ describes this volume, P describes an identical volume obtained in translating $V_{ij}$ by vector $\overrightarrow{B_{ij}P}$, which is constant because the orientation is fixed. This volume noted $V_{ij}^{p}$ is the working volume allowed for P under the constraints on rod ij. The workspace being the one where the constraints on all rods are satisfied, it is obtained by the intersection of all volumes $V_{ij}^{p}$. This volume is noted $V^{p}$ [1].

To facilitate the calculation of the workspace, instead of determining the volume $V_{ij}$ reachable by the point $B_{ij}$, we express the coordinates of $B_{ij}$ as function of the coordinates of P and we look directly for the working volume $V_{ij}^{p}$ allowed for P. The constraints that limit the workspace of a parallel manipulator are (i) Interference between links, (ii) Leg Length, (iii) Serial Singularity, (iv) Mechanical limits on passive joints and (v) Actuator stroke.

In the following three subsections, we model geometrically these constraints under the hypothesis that $\rho_{11}$ can be different from $\rho_{12}$. Then we consider in subsection F the interdependence between rods 11 and 12 of leg I characterized by the fact that $\rho_{11} = \rho_{12} = \rho_{1}$.

### C.  Interference between the various elements of the machine

In this subsection, we define a parallelepiped rectangle for point P assuring that there is no collision between the various elements of the VERNE machine. In order to do that, we represent the corresponding constraints by inequalities that allow us to determine the intervals of $x_p$, $y_p$ and $z_p$.

To avoid slider-leg collisions, the leg should be only in one of the half-spaces separated by the plane parallel to the slider face and passing through point $A_{ij}$ (see Fig. 2). From an analytical point of view, we can represent this constraint by inequalities that allow us to determine the interval $]x_{p\min}, x_{p\max}[$ of $x_p$ as shown in Eqs. (7a) and (7b).

$$x_{Bij} - x_{Aij} < 0 \ (i = 1, j = 1..2) \iff x_{p\max} = d_1 - D_1 \tag{7a}$$

$$x_{Bij} - x_{Aij} > 0 \ (i = 2..3, j = 1) \iff x_{p\min} = d_2 - D_2 \tag{7b}$$

To determine the interval $]y_{p\min}, y_{p\max}[$, we used the leg-length constraint:

$$y_{Bij} - y_{Aij} < L_2 \ (i = 2, j = 1..2) \iff y_p < R_2 \cos(\alpha) + L_2 - r_4 \tag{8a}$$

$$y_{Aij} - y_{Bij} < L_3 \ (i = 3, j = 1..2) \iff y_p > r_4 - R_2 \cos(\alpha) - L_3 \tag{8b}$$

As our goal is only to find, for any value of $\alpha$, the minimal interval $]y_{p\min}, y_{p\max}[$, we use $\alpha = 0$ in Eqs (8a) and (8b), which implies that $y_{p\max} = L_2 + R_2 - r_4$ and $y_{p\min} = r_4 - R_2 - L_3$.

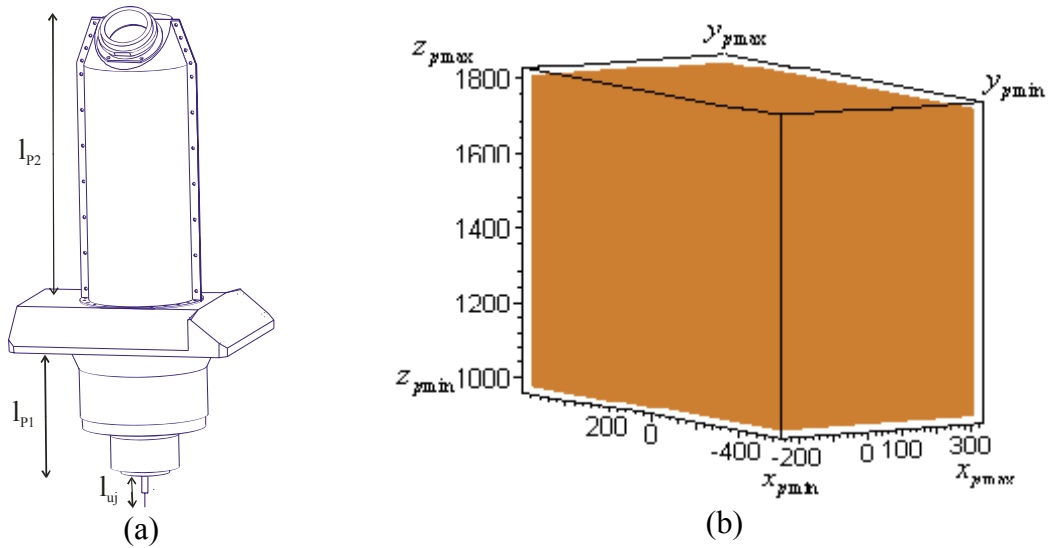

Figure 5: (a) Platform body and (b) the allowable volume for point P assuring that there is no collision between the various elements of the machine and for a tool length l=0.

To evaluate the performance of the machine, it is important to find the workspace of the machine for various tool lengths. Let us define $d_{p\_u} = l_{p1} + l_{uj}$ as the distance between the platform and the extremity U of the tool (Fig. 5a). In order that there is no collision between the tool and the plate, it is necessary that at the limit the tool nears the tilting table.

$$z_U - z_{tilting\ table} < 0 \Rightarrow z_P < z_{tilting\ table} - d_{p\_u} \cos(\alpha) \tag{9a}$$

Let us take $\alpha = 0$ in Eq. (9a), $z_{P\max} = z_{tilting\ table} - d_{p\_u}$. This condition insures that there is no collision between the platform and the tilting table for any value of $\alpha$.

To insure that there is no interference between the platform and the hood of the machine, we define this constraint:

$$z_P > z_{hood} + l_{p2} \cos(\alpha) \tag{9b}$$

Let us take $\alpha = 0$ in Eq. (9b), $z_{P\min} = l_{p2} + z_{hood}$. This condition insures that there is no collision between the platform and the hood of the machine for any value of $\alpha$

Knowing the lower and superior boundary of $x_p$, $y_p$ and $z_p$, we can build in the fixed Cartesian frame the parallelepiped rectangle ${}^1V^P$, which represents the domain where there is no risk of collision between the different elements of the Verne machine (see Fig. 5b).

### D. Leg length limits and Serial singularity constraint

To simplify the explanation of our method, we define points $A'_{ij}$ obtained by translating points $A_{ij}$ by vector $\overrightarrow{B_{ij}P}$, which is constant because the orientation $\alpha$ is fixed. Points $A_{ij}$ (respectively $A'_{ij}$) correspond to the real centers (respectively imaginary centers) of rotation of the joints connected to the base. The introduction of $A'_{ij}$ will allow us to find directly the volume ${}^kV^P_{ij}$ allowed by P under the constraint k on rod ij (we suppose that the constant orientation workspace of the platform is a volume because leg I is virtually cut).

Let the length of a leg be $L_i$, if we respectively indicate by $A'_{ij,0}$ and $A'_{ij,1}$ the initial and final position of the center of joint ij, thus when point $A'_{ij}$ coincides with point $A'_{ij,0}$, the set of points reachable by P is a sphere $S'_{ij,0}$ of radius $L_i$ and of center $A'_{ij,0}$. However, a particular characteristics of parallel manipulators of type $\underline{P}SS$ (or $\underline{P}US$) is that they undergo a serial singularity at configurations where a rod ij is perpendicular to its corresponding guideway axis. Since passing through such a singularity is undesirable, the motion of each rod should be restricted so that the angle between the vectors $\overrightarrow{A'_{ij}B'_{ij}}$ and $\overrightarrow{A'_{ij,0}A'_{ij,1}}$ be always in only one of the two ranges $\left[0, \dfrac{\pi}{2}\right[$ or $\left]\dfrac{\pi}{2}, \pi\right]$. So when the actuator is in its initial position $A'_{ij,0}$, we can define for every rod ij a plane perpendicular to axes $(A'_{ij,0} \ A'_{ij,1})$ at $A'_{ij,0}$. This plane divides the sphere $S'_{ij,0}$ into two hemispheres. We take the hemisphere that is on the side of the axis $(A'_{ij,0} \ A'_{ij,1})$. When the slider reaches its final position by starting from $A'_{ij,0}$, the volume described by P, noted $^2V_{ij}^P$, is the volume swept by the chosen hemisphere along $(A'_{ij,0}, A'_{ij,1})$ (Fig. 6).

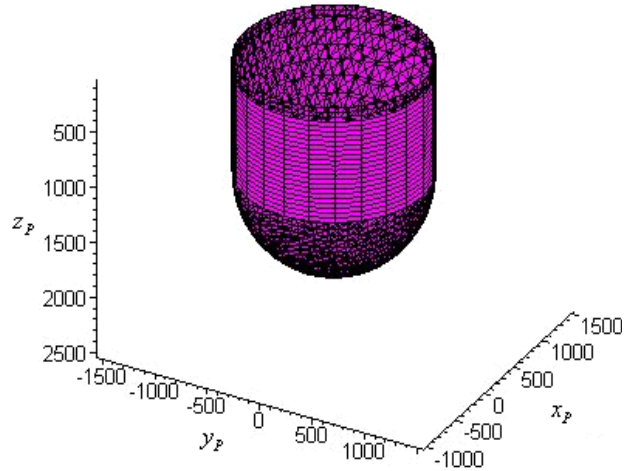

Figure 6: Workspace of the point P under the constraints imposed by length of the rod ij, without singularity and for a given orientation

The volume $^2V_{ij}^P$ represents the locations of point P under the constraints on the length of the rod ij and by making sure that rod ij does not pass through a serial singularity. The volume $^2V^P = {}^2V_{11}^P \cap {}^2V_{12}^P \cap {}^2V_{21}^P \cap {}^2V_{31}^P$ satisfies the constraints imposed by leg lengths and does not contain any serial singularity. This volume is obtained for a given orientation $\alpha$ and by considering that the leg I is virtually cut, so other serial singularities can be found for the real Verne machine when we consider the interdependence between rods 11 and 12 of leg I. This singularity will be presented in the following subsection F.

### E. Mechanical limits on passive joints

In this subsection, we propose a method that allows taking into account mechanical limits on the passive joints. This method takes into consideration the type of joints as well as the location of these joints in the machine. Therefore, our goal is to find geometric models of these constraints to be able to find the topology of the boundary of the workspace. In this subsection, we suppose that the constant orientation workspace of the platform is a volume because leg I is virtually cut.

We have to mention that point P is obtained by translating $B_{ij}$ by vector $\overrightarrow{B_{ij}P}$, which is constant because the orientation is fixed. Because our purpose is to find directly the locations of P, we suppose that rod ij is linked to the corresponding prismatic joint and to the moving platform by points $A_{ij}^{'}$ and $P$, respectively, instead of points $A_{ij}$ and $B_{ij}$, respectively.

*1) Mechanical limits on the passive joints attached to the prismatic joints*

We can describe the movement of the spherical joints by 3 angles $\beta$, $\delta$, $\gamma$. These angles are defined in the following way: starting from the frame $R_{ij,0} = (A_{ij}^{'}, x_{ij,0}, y_{ij,0}, z_{ij,0})$ linked to the joint ij, we obtain the orientation of the base joint ij by a first rotation about the y-axis through an angle $\delta$ then by rotating about the new $x_{ij,1}$ axis by an angle $\beta$ and finally by rotating about the new z-axis through an angle $\gamma$. However, the rotation of the rod about the last axis can be ignored because there are no joint limits. The obtained final frame is $R_{ij,2} = (A_{ij}^{'}, x_{ij,2}, y_{ij,2}, z_{ij,2})$ (Fig. 7).

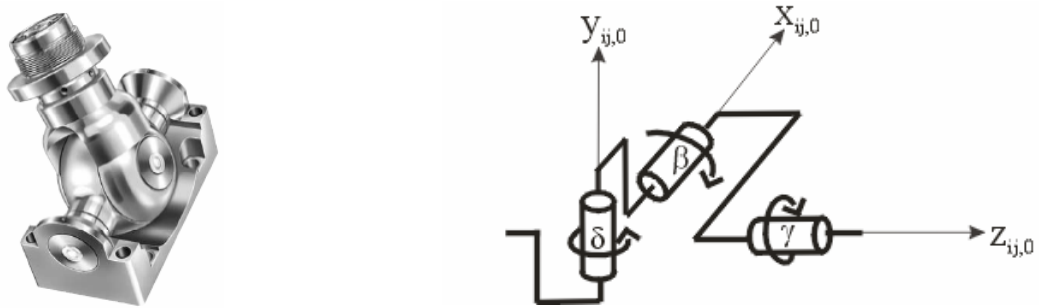

Figure 7: Spherical joint

Since designers provide graphs that describe the movement of joints (Fig. 8), we can define the angle $\beta$ as function of $\delta$. So knowing the angular ranges of these angles as well as the function that relates them (Fig 8), we can describe the motion of these joints.

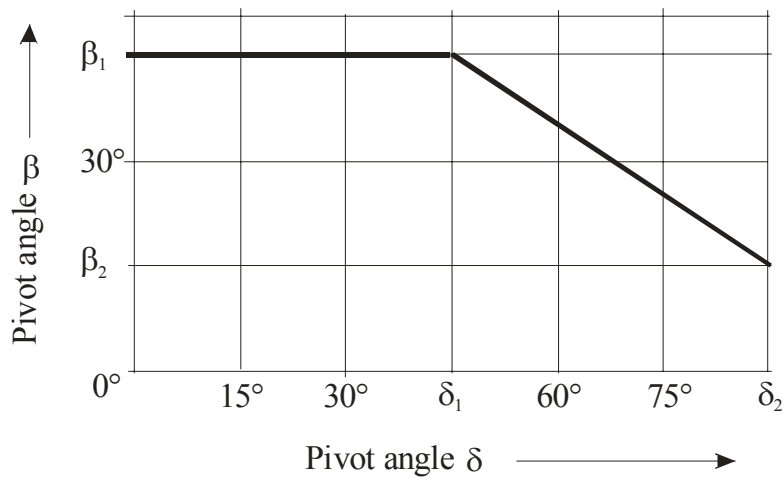

Figure 8: Definition of mechanical limits of a joint of type INA GLK 3

The modeling of the constraints due to the mechanical limits on the base joints

Let us $^{ij,0}T_{ij,2} = \begin{bmatrix} ^{ij,0}s_{ij,2} & ^{ij,0}n_{ij,2} & ^{ij,0}a_{ij,2} & 0 \end{bmatrix} = Rot(y_{ij,0},\delta)Rot(x_{ij,1},\beta)$ define the transformation between frame $R_{ij,0}$ and frame $R_{ij,2}$ where, $^{ij}s_{ij,2}$, $^{ij}n_{ij,2}$ and $^{ij}a_{ij,2}$ indicate respectively the unit vectors of axes $x_{ij,2}$, $y_{ij,2}$ and $z_{ij,2}$ of $R_{ij,2}$ expressed in $R_{ij,0}$ and where $Rot(y_{ij,0},\delta)$ (respectively $Rot(x_{ij,1},\beta)$) is a rotation matrix describing the rotation of angle $\delta$ (respectively $\beta$) $y_{ij,0}$ (respectively $x_{ij,1}$).

$$^{ij,0}T_{ij,2} = Rot(y_{ij,0},\delta)Rot(x_{ij,1},\beta) = \begin{bmatrix} C\delta & 0 & S\delta & 0 \\ 0 & 1 & 0 & 0 \\ -S\delta & 0 & C\delta & 0 \\ 0 & 0 & 0 & 1 \end{bmatrix}\begin{bmatrix} 1 & 0 & 0 & 0 \\ 0 & C\beta & -S\beta & 0 \\ 0 & S\beta & C\beta & 0 \\ 0 & 0 & 0 & 1 \end{bmatrix} \qquad (10)$$

where $S\delta$ and $C\delta$ denote $\sin\delta$ and $\cos\delta$, respectively.

Let vectors $^{ij,0}P$ and $^{ij,2}P$ define the expression of P in frames $(A_{ij}^{'}, x_{ij,2}, y_{ij,2}, z_{ij,2})$ and $(A_{ij}^{'}, x_{ij,0}, y_{ij,0}, z_{ij,0})$, respectively. Knowing $^{ij,2}P = \begin{bmatrix} 0 & 0 & L_i & 1 \end{bmatrix}^T$ and $^{ij,0}T_{ij,2}$ from Eq. (10) we deduce $^{ij,0}P$:

$$^{ij,0}P = {}^{ij,0}T_{ij,2}\,{}^{ij,2}P = \begin{bmatrix} L_i\sin(\delta)\cos\beta & -L_i\sin\beta & L_i\cos(\delta)\cos\beta & 1 \end{bmatrix}^T \qquad (11)$$

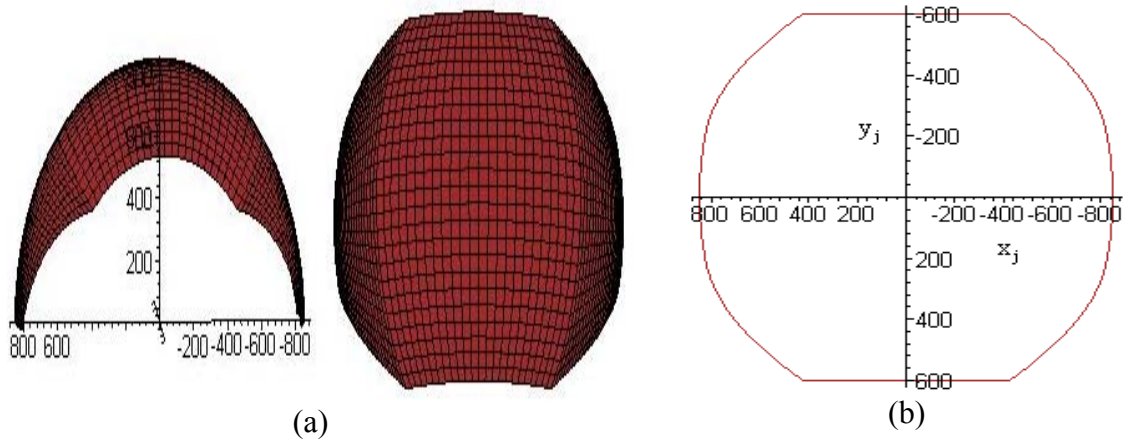

(a)                                                                (b)

Figure 9 (a) Definition of the surface characterizing constraints on the base joint for a fixed position of the corresponding slider. The point $A_{ij}^{'}$ is fixed; the point P describes this constraint in the frame linked to the joint. (b) Projection of this surface in the plane xOy

Since the coordinates of P are expressed as function of angles $\beta$ and $\delta$ (Eq. 11) and $\beta$ is expressed as function of $\delta$, it is sufficient to know the angular range of $\delta$ (Fig 8) to define geometrically the locations of P in the frame linked to the corresponding joint (Fig. 9).

<u>Location of the universal joints</u>

Let us consider the frame $R_{ij,0} = (A'_{ij}, x_{ij,0}, y_{ij,0}, z_{ij,0})$ linked to joint ij and the fixed Cartesian frame $R_a = (O, x, y, z)$. Let us define $^a T_{ij,0} = \begin{bmatrix} ^a s_{ij,0} & ^a n_{ij,0} & ^a a_{ij,0} & ^a H_{ij,0} \end{bmatrix}$, the transformation that brings the frame $R_a$ on the frame $R_{ij,0}$ where $^a H_{ij,0}$ is the vector expressing $A'_{ij}$, origin of the frame $R_{ij,0}$ in the frame $R_a$. Let us denote $\theta_{ij}$, $\psi_{ij}$, $\phi_{ij}$, the angles defining the position of the universal joint ij with respect to $R_a$. These angles are defined in the following way: starting from the fixed Cartesian frame $R_a$, we obtain the frame $R_{ij,0}$ linked to the joint ij by undergoing a translation along the vector $\overrightarrow{OA'_{ij}}$, then by rotating about the y-axis through an angle $\psi$, then by rotating about the new x-axis through an angle $\theta$ and finally by rotating about the new z-axis through an angle $\phi$.

$$^a T_{ij,0} = Trans(x_{A'_{ij}}, y_{A'_{ij}}, z_{A'_{ij}}) Rot(y, \psi_{ij}) Rot(x, \theta_{ij}) Rot(z, \phi_{ij}) \tag{12a}$$

where $Rot(y, \psi_{ij})$, $Rot(x, \theta_{ij})$ and $Rot(z, \phi_{ij})$ are respectively the rotation matrixes describing the rotation of angles $\psi_{ij}$, $\theta_{ij}$ and $\phi_{ij}$ about the y, x and z axis.

$$^a T_{ij,0} = \begin{bmatrix} C\psi_{ij}C\phi_{ij} + S\psi_{ij}S\theta_{ij}S\phi_{ij} & S\psi_{ij}S\theta_{ij}C\phi_{ij} - C\psi_{ij}S\phi_{ij} & S\psi_{ij}C\theta_{ij} & x_{A'_{ij}} \\ C\theta_{ij}S\phi_{ij} & C\theta_{ij}C\phi_{ij} & -S\theta_{ij} & y_{A'_{ij}} \\ C\psi_{ij}S\phi_{ij}S\theta_{ij} - S\psi_{ij}C\phi_{ij} & C\psi_{ij}S\theta_{ij}C\phi_{ij} + S\psi_{ij}S\phi_{ij} & C\psi_{ij}C\theta_{ij} & z_{A'_{ij}} \\ 0 & 0 & 0 & 1 \end{bmatrix} \tag{12b}$$

Let vectors $^a P$ and $^{ij,0}P$ represent the point P respectively expressed in frames $R_a$ and $R_{ij,0}$ as function of angles $\delta$, $\beta$. Knowing $^{ij,0}P$ from Eq. (11) and $^{ij,0}T_a$ from Eq. (12), we deduce:

$$^a P = {}^a T_{ij,0} \; {}^{ij,0}P \tag{13}$$

The matrix $^a T_{ij,0}$ is expressed as function of $\rho_i$, which depends on the position of the slider, and as function of angles $\theta_{ij}$, $\psi_{ij}$, $\phi_{ij}$ and $\alpha$ (the coordinates of $A'_{ij}$ are expressed as function of $\rho_i$ and $\alpha$). The first three angles design parameters and the last one is the angle representing the orientation of the platform, which is assumed known. Thus $^a P$ can be expressed as function of $\beta$, $\delta$ (described in Fig 8) and $\rho_i$ only. We can now define the acceptable spherical region $S'_{ci}$ for P when $A'_{ij}$ is fixed (Fig. 10), by varying $\delta$ from $-\delta_2$ to $+\delta_2$ in Eq. (13) (remember $\beta$ is expressed as function of $\delta$ as shown in Fig 8). Since $A'_{ij}$ can only move between points $A'_{ij,0}$ and $A'_{ij,1}$, the acceptable spatial region for P is the volume $^3V^P_{ij}$ swept by $S'_{ci,0}$ along $A'_{ij,0} A'_{ij,1}$. This volume is the volume allowed for P by taking into account only mechanical limits on the base joint ij. The total volume including all base joints will be: $^3V^P = {}^3V^P_{11} \cap {}^3V^P_{12} \cap {}^3V^P_{21} \cap {}^3V^P_{31}$.

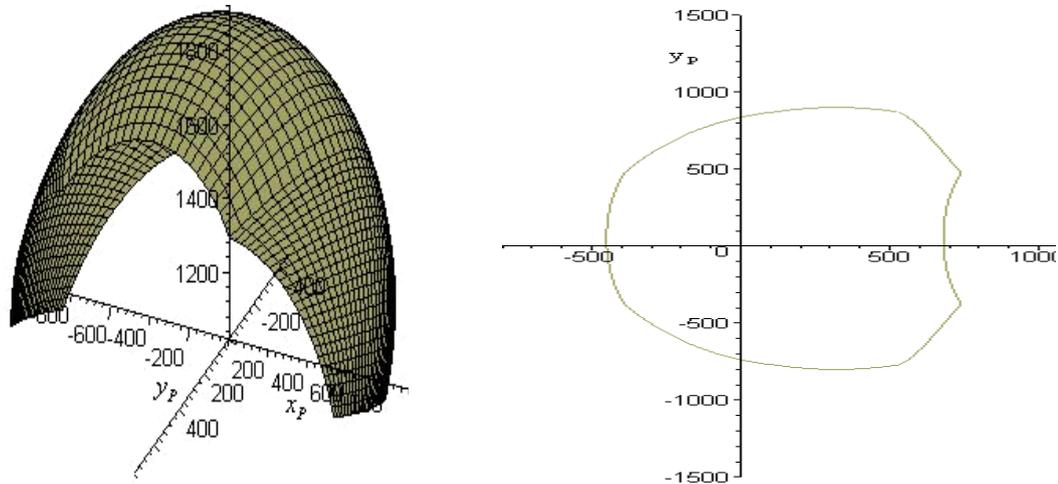

Figure 10: (a) Locations of point P characterizing the constraints on the base joint ij, for a given orientation and for a fixed position of the corresponding slider. (b) Projection of this surface in the plane xOy

In order to obtain the projection of $S'_{ci,0}$ in the plane xOy, we calculate the x and y coordinates of $^aP$ from Eq. (13) by varying $\delta$ between $\max(-\delta_2, -\pi/2 - \psi_{ij})$ and $\min(\delta_2, \pi/2 - \psi_{ij})$, in this way, we also insure that the leg ij does not pass through a serial singularity, which occurs when the angle between the corresponding slider and the leg is higher than $\pi/2$.

### 2) Mechanical limits on the passive joints linked to the moving platform

In order to model constraints imposed by the platform joints, we can clearly adopt the same model as the one used for the joints attached to the prismatic joints. So we can define the spherical region $S'_{si}$ (we use the same reasoning as in the previous subsection) of vertex P such as if the constraints on the joint ij are satisfied, the point $A'_{ij}$ is on this surface. Let us consider the reference frame $(A'_{ij}, x'_{ij,2}, y'_{ij,2}, z'_{ij,2})$ where $x'_{ij,2} = -x_{ij,2}$, $y'_{ij,2} = -y_{ij,2}$ and $z'_{ij,2} = -z_{ij,2}$. We can define in this reference frame a surface equivalent to $S'_{si}$, $S'^P_{si}$ of vertex $A'_{ij}$ such as if the constraints on the platform joint ij satisfied then the point P is on surfaces $S'^P_{si}$ [13]. Since the point $A'_{ij}$ can only move between points $A'_{ij,0}$ and $A'_{ij,1}$, thus the acceptable spatial region for the point P is the volume $^4V^P_{ij}$ swept by $S'^P_{si,0}$ along $A'_{ij,0} A'_{ij,1}$. This volume is the volume allowed for P by taking into account only mechanical limits on the platform joint ij, the total volume including all platform joints will be: $^4V_i^P = {}^4V_{11}^P \cap {}^4V_{12}^P \cap {}^4V_{21}^P \cap {}^4V_{31}^P$

### F. Closure constraints

We consider the interdependence between the rods of leg I that allows us to define the region in which P can move. This region is a surface.

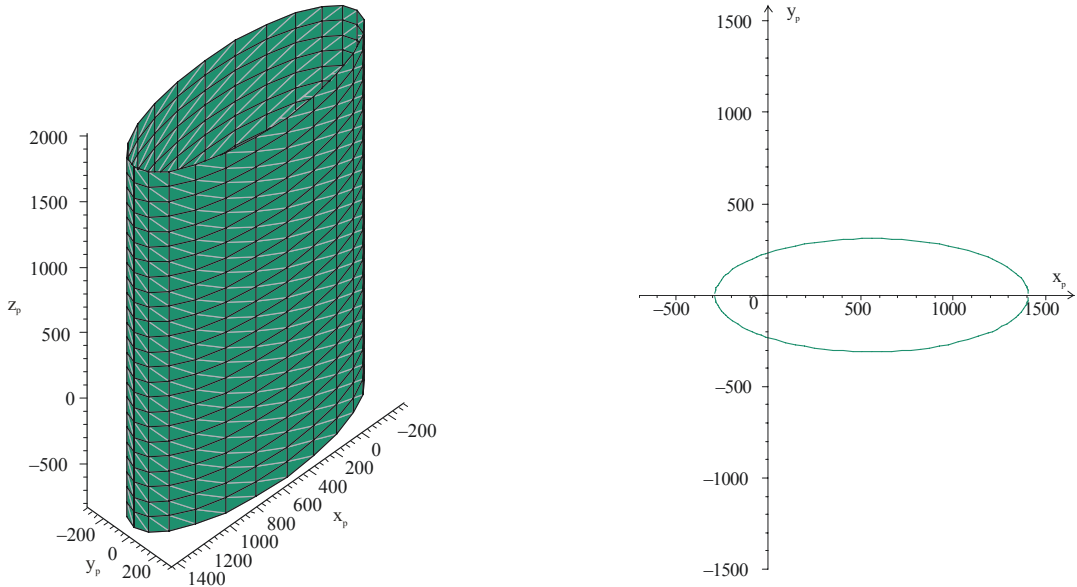

Figure 11: (a) Locations of point P characterizing the constraint imposed by the shape of the leg I for a given orientation $\alpha$ (b) The projection of this surface in the plane xOy

For a given orientation $\alpha$, we can model the Eq. (6) by a hollow cylinder $Cy_1^P$ whose base is an ellipse (Fig. 11). This surface represents the coupling between the position and the orientation of the platform resulting from the shape of the Leg I. However the serial singularities and the inverse kinematics induce additional constraints. In the previous subsection we studied the serial singularities for the VERNE parallel module in the virtual shape that is when the leg I is cut. However, we have conducted previous studies in order to find the serial singularities for the VERNE parallel module [14]. In this study, we demonstrated that a particular singularity was obtained when $R_1 \cos(\alpha) = r_1$, so to avoid passing by this singularity it is necessary to impose the following constraint:

$$\cos(\alpha) > \frac{r_1}{R_1} \qquad (14)$$

For the workspace calculation of the VERNE machine only one solution to the inverse kinematic problems is taken into consideration. In a previous study [14], we have proved that this solution is obtained when signs of $\rho_1 - z_P$, $\rho_2 - z_P + R_2 \sin(\alpha)$ and $\rho_3 - z_P - R_2 \sin(\alpha)$ are negative. However subtracting equation (2a) from equation (2b) yields:

$$y_P \left( R_1\cos(\alpha) - r_1 \right) = R_1\sin(\alpha)\left( \rho_1 - z_P \right) \qquad (15)$$

Equation (15) implies that $\text{sgn}\left( \rho_1 - z_P \right)\text{sgn}(\sin(\alpha)) = \text{sgn}(R_1\cos(\alpha) - r_1)\text{sgn}(y_P)$. So in order to verify the inverse kinematic condition $\rho_1 < z_P$ and the serial singularity condition $R_1\cos(\alpha) > r_1$ only half of the hollow cylinder $Cy_1^P$ obtained for $y_P < 0$ (respectively $y_P > 0$) if $\alpha > 0$ (respectively $\alpha < 0$) must be considered in the workspace calculation of the VERNE machine.

It is important to mention that the existence of this constraint directed us to look for the locations of the point P in a plane parallel to the plane $(xOy)$ for a given orientation $\alpha$.

### G. The Complete Workspace Algorithm

Until now, we looked for the workspace of the VERNE parallel module by considering each constraint separately. In this subsection, we first suppose that the orientation of the moving platform is fixed and we find all the possible positions of the point P verifying all constraints in a horizontal plane. Then we scan parameters $\alpha$ and $z_P$ to find the complete workspace.

The workspace of the VERNE parallel module varies with $z_P$. However we can prove graphically that for $z_P \in I_2 = \left[ \max(\rho_{i\min} + l_i),\ \min(\rho_{i\max}) \right]$ $(i = 1..3)$, the workspace remains constant. When $z_P \in I_1 \cup I_3 = \left[ \min(\rho_{i\max}),\ \min(\rho_{i\max} + l_i) \right] \cup \left[ \max(\rho_{i\min}),\ \max(\rho_{i\min} + l_i) \right]$ $(i = 1..3)$, the workspace varies as shown in Fig. 12 where $\rho_{i\min}$ and $\rho_{i\max}$ represent the minimal and the maximal value of $\rho_1$, respectively.

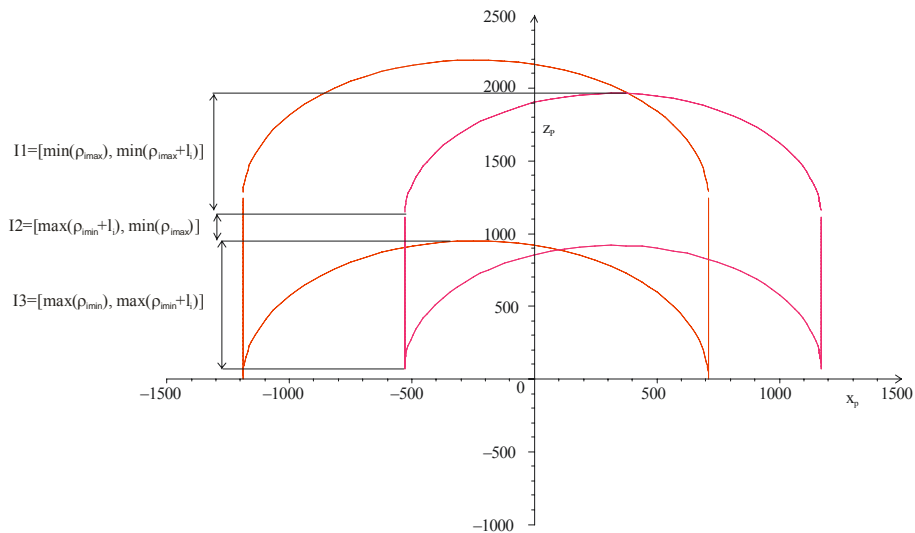

Figure 12: Projection of the volume $^2V^P$ in the plane xOz for a given orientation $\alpha$

Summary of the method

The workspace of the VERNE is the intersection between geometric models already defined in the previous section. To calculate this workspace, we apply the following steps:

*Step1:   We vary $\alpha$ between $-\alpha_1$ and $+\alpha_1$ following a constant step and for each value of $\alpha$ we repeat Step2, Step 3 and Step4.*

$$\alpha_1 = \arccos\left(\frac{r_1}{R_1}\right) \tag{16}$$

*Step2:   We horizontally project all geometrical models obtained for a given orientation $\alpha$ (see Fig. 13)*

*Step3   We calculate the intersection between the projected zones characterizing leg length constraints, interference between links and mechanical limits on passive joints to obtain the zone $S_1(\alpha)$ (see Fig. 14).*

*Step4:   We select the part of half of the ellipse that is inside the zone $S_1(\alpha)$.*

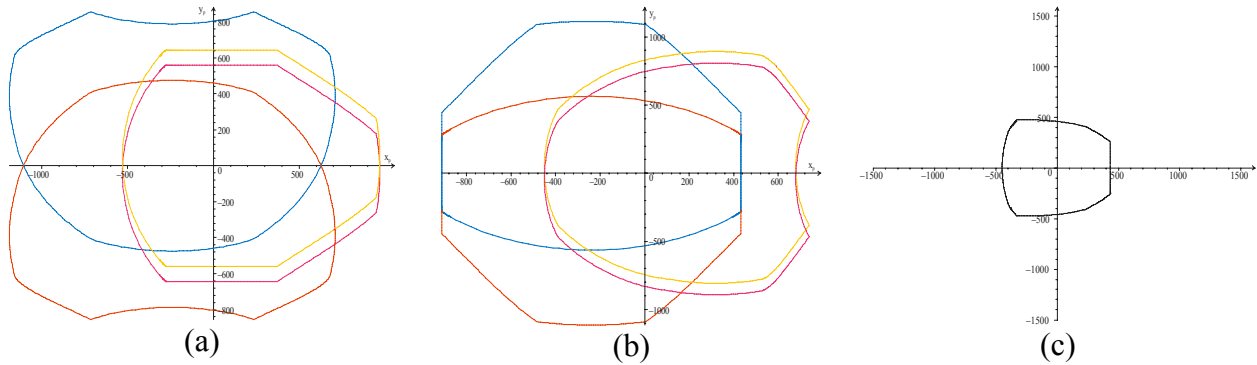

Figure 13: Locations allowed for P in the horizontal plane and for a given orientation $\alpha$ ; (a) under the base joint mechanical limits, (b) under the platform joint mechanical limits, (c) by considering all passive joint constraints (after intersection between curves)

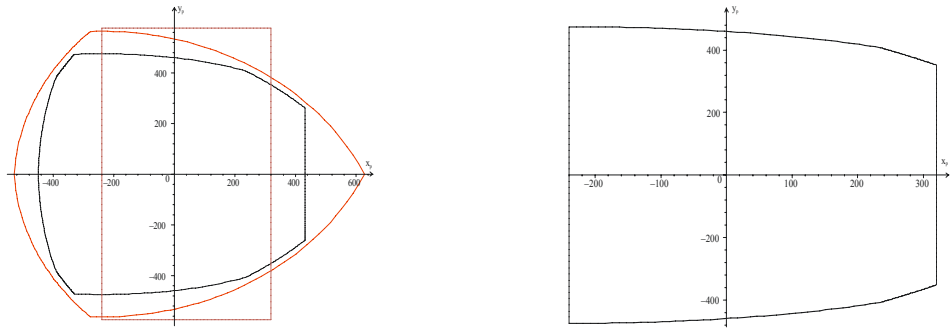

Figure 14: Zone $S_1(\alpha)$ before and after intersection between curves characterizing Leg length constraints, mechanical limits on passive joint and interference between links

*Step5:   We take discrete values of $z_P$ in the interval $\left[(z_{hood} + l_{p2}, z_{tilting\ table} - d_{p\_u}\right]$ and for each value of $z_P$ we apply Step6 to Step10.*

*Step6: We vary $\alpha$ between $-\alpha_1$ and $+\alpha_1$ following a constant step and for each value of $\alpha$ we repeat the following steps from Step7 to Step10.*

*Step7: If $z_P \in I_2$, we use results from Step4 and we go to Step10 else we go to Step8.*

*Step8: If $z_P \in I_1 \cup I_3$, we horizontally cut all volumes ${}^2V_k^P$ ($k = 11, 12, 21, 31$) by the plane $z = z_P$ and then we calculate the intersection between the obtained circular zones and the zone $S_1(\alpha)$ to get a new zone $S_2(\alpha, z_P)$.*

*Step9: If $z_P \in I_1$, we select the part of half of the ellipse, which is inside the zone $S_2(\alpha, z_P)$. Other wise if $z_P \in I_3$, we select the part of half of the ellipse, which is outside the zone $S_2(\alpha, z_P)$ with respect to $S_1(\alpha)$. Example is illustrated in Fig. 15.*

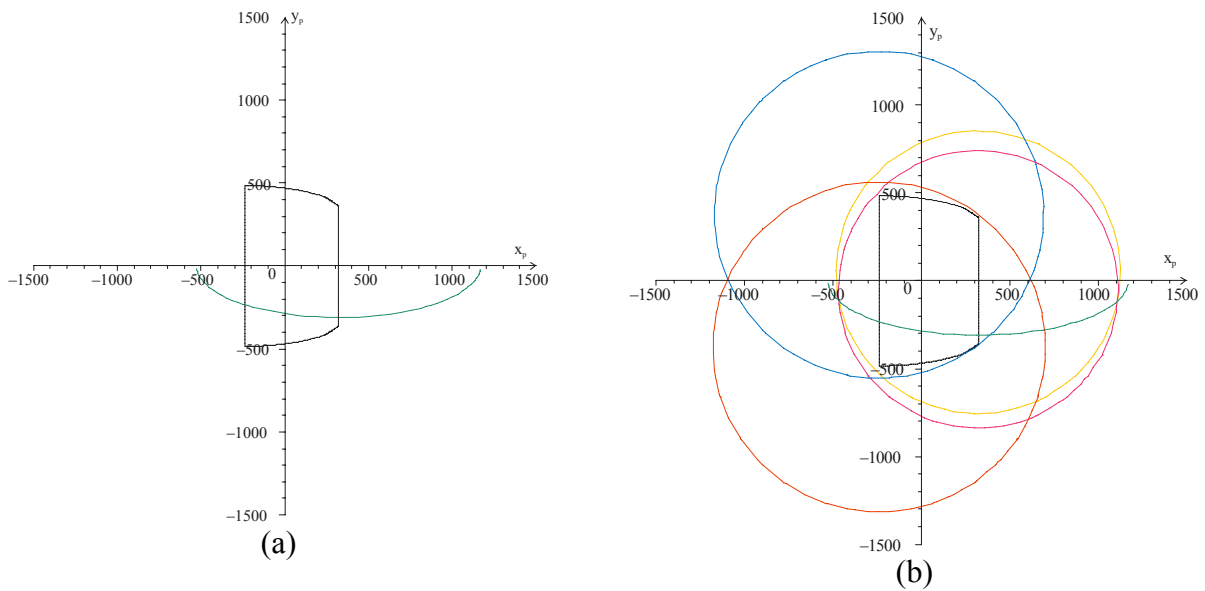

(a)                                    (b)

Figure 15: The places of the point P for a given orientation and height with (a) $z_P \in I_2$, (b) $z_P \in I_1 \cup I_3$

*Step10: We have $y_P$ as function of $x_P$ (Eq. (5)) and we know their lower and upper bounds for a given $z_P$. Thus we first save these values in a note pad file $f1$ then we vary $x_P$ following a constant step and we calculate $y_P$ for each value of $x_P$. This will permit us to save numerically all values of $x_P$, $y_P$ and $z_P$ belonging to the workspace in another note pad file $f2$.*

*Step11: In order to obtain the volume of the workspace, we import the file $f1$ into CATIA V5® and we make the meshing by using "Quick surface reconstruction product". After that, we recover the surface by using the "Digitized Shape Editor product". Finally we fill the surface using the fill function in the "part design product". The obtained volume is the complete workspace as shown in Fig. 16.*

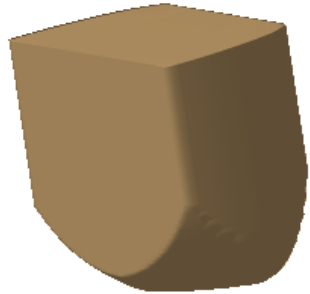

Figure 16: Workspace defined as the set of points P in the fixed Cartesian frame $R_b$

Now in order to calculate the workspace for a given tool length we proceed as follows:

*Step12: We express the coordinates of the tool centre point Ou $(x_{to}, y_{to}, z_{to})$ in the fixed Cartesian frame as function of the coordinates of the point P*

$$x_{to} = x_P, \quad y_{to} = y_P - d_{p\_u}\sin(\alpha) \text{ and } z_{to} = z_P + d_{p\_u}\cos(\alpha) \tag{17}$$

Where $d_{p\_u} = l_{p1} + l_{uj}$ and $l_{uj}$ is the tool length (Fig. 5a).

*Step13: We add the condition that $z_{to} \leq z_{tilting\ table}$*

*Step14: We save the obtained results in a note pad file.*

*Step15: We do the same operations as for Step 11. This will permit us to obtain a volume that represents the complete workspace for a given tool length as shown in Fig. 17.*

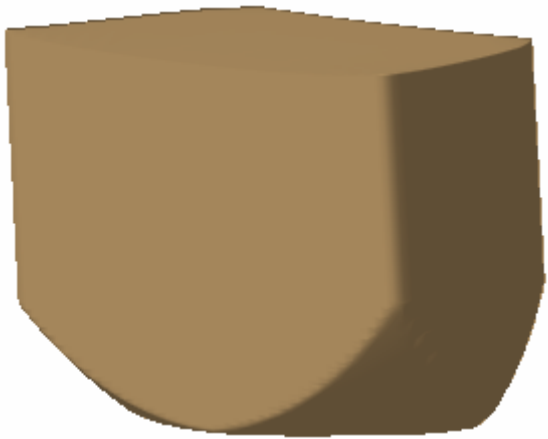

$L_{u1} = 50mm$

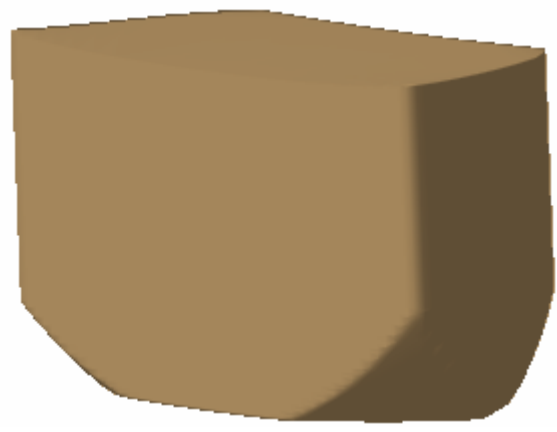

$L_{u2} = 100mm$

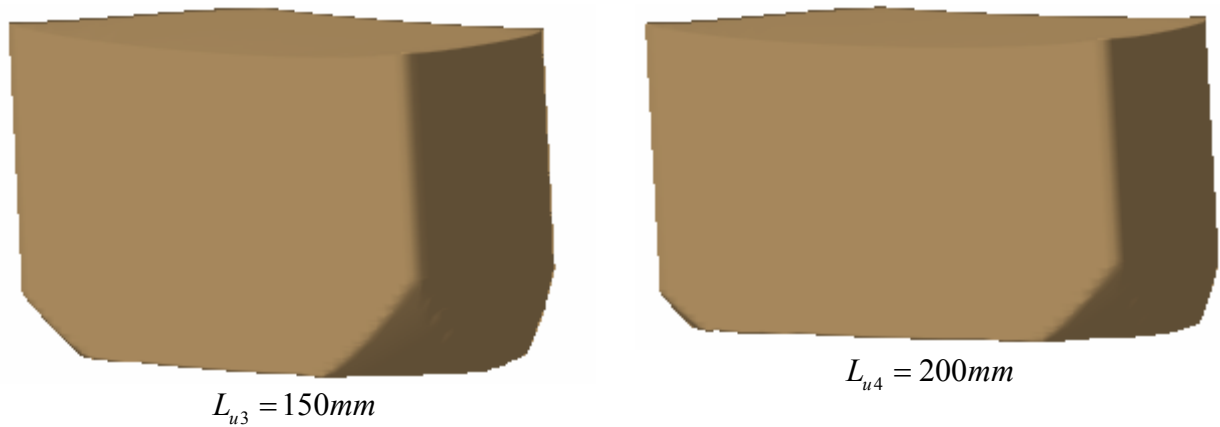

$L_{u3} = 150mm$    $L_{u4} = 200mm$

Figure 17: The workspace for various tool lengths

The proposed procedure for calculating the workspace of the VERNE parallel module for various tool lengths was implemented in Maple 10®.

## IV. CONCLUSIONS

In this paper, we have proposed a method for calculating the complete workspace for various tool lengths of the VERNE parallel module. This method takes into account all constraints having an actual influence on the workspace of the VERNE: leg length, serial singularity, mechanical limits on passive joints, actuator stroke, interference between links and closure constraints. The last constraint was a particular one for the VERNE due to the shape of the leg I which is not a parallelogram.

In this method, leg I was cut by considering that it is linked to the base through two prismatic joints instead of one and we geometrically modeled constraints limiting the workspace of the new parallel architecture for a given orientation of the platform. The intersection between these models is a volume. We geometrically modeled the closure constraint for a given orientation of the platform; this gave us a surface. We calculated the intersection between these models in a known horizontal plane, then we proceeded by discretization to determine the complete workspace of the VERNE parallel module. An algorithm for the determination of the workspaces was also presented. This algorithm was implemented in the computer algebra Maple 10®. Examples were provided to illustrate the results.

This method was particularly applied for the Verne machine. However it can be always applied to all machines of type PSS (PUS or PSU) but with some modifications in the algorithm according to the shape of legs and number of degree of freedom of the machine. For parallel manipulators with more than three degrees of freedom, we can fix angles representing the orientation degrees of freedom and then calculate a constant orientation workspace. For parallel manipulators where legs are simple rods (or of parallelogram shape) we can
geometrically model constraints and then directly implement these models into CATIA V5® to
be able to calculate the intersection between these models and obtain the workspace volume.

ACKNOWLEDGMENTS

This work has been partially supported by the European projects NEXT, acronyms for "Next
Generation of Productions Systems", Project no° IP 011815. The authors would also like to
thank the Fatronik society, which permitted us to use the CAD drawing of the Machine VERNE
what allowed us to present well the machine.